\documentclass{article}

\usepackage{microtype}
\usepackage{graphicx}
\usepackage{booktabs}
\usepackage{hyperref}

\usepackage[accepted]{icml2026}

\usepackage{amsmath}
\usepackage{amssymb}

\usepackage[capitalize,noabbrev]{cleveref}

\usepackage[disable]{todonotes}

\usepackage{tikz}
\usetikzlibrary{positioning,arrows.meta,shapes.geometric,shapes.symbols,calc}

\icmltitlerunning{Compressing the Validation Bottleneck}

\begin{document}

\twocolumn[
\icmltitle{Compressing the Validation Bottleneck: \\
An Agentic Self-Driving Lab for Scientific Discovery}

\icmlsetsymbol{equal}{*}

\begin{icmlauthorlist}
\icmlauthor{Kyunghoon Hur}{keti}
\icmlauthor{Chihun Lee}{kims}
\end{icmlauthorlist}

\icmlaffiliation{keti}{AX Research Division, Korea Electronics Technology Institute, Republic of Korea}
\icmlaffiliation{kims}{Material Data Division, Korea Institute of Materials Science, Republic of Korea}
\icmlcorrespondingauthor{Kyunghoon Hur}{kyunghoonhur@keti.re.kr}
\icmlcorrespondingauthor{Chihun Lee}{chihunlee@kims.re.kr}

\icmlkeywords{Self-Driving Labs, Agentic AI, Lab-in-the-Loop, Design of Experiment, Surrogate Characterization, Prior-Aware Agent, Cost-Aware Agent, Bayesian Optimization}

\vskip 0.3in
]

\printAffiliationsAndNotice{}

\begin{abstract}
Agentic AI-for-Science can automate ideation, planning, and analysis, but final validation still depends on real experiments. A self-driving lab (SDL) can execute those experiments, yet the loop still has bottlenecks: the agent may spend too many rounds on low-value experiments, or each round may require a high-cost experiment. We target these two physical bottlenecks with one agent. First, a prior-aware agentic DOE loop uses domain knowledge and past results to propose feasible and informative next experiments, reducing trials-to-target. Second, a cost-aware surrogate agent predicts high-cost, high-resolution measurements from low-cost, low-resolution measurements. It chooses between a high- and a low-cost measurement based on the predicted uncertainty. We examine these directions in the biology and materials domains, respectively. Together, under a single agent, these components aim to accelerate the SDL loop by reducing both the number of loops and the cost per experiment.
\end{abstract}

\begin{figure*}[t]
\centering
\includegraphics[width=0.95\textwidth]{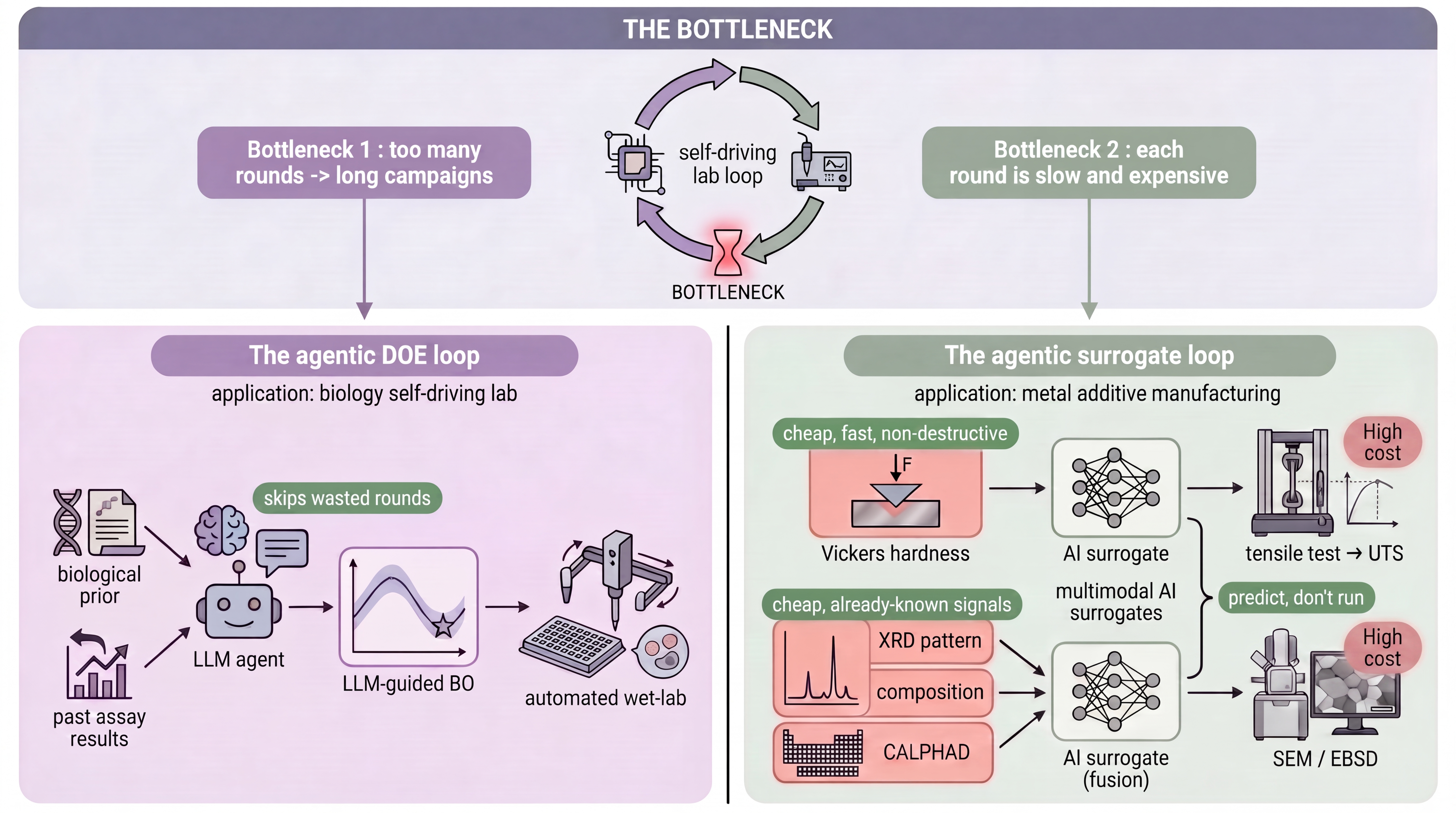}
\caption{One agent attacks two bottlenecks of a domain-general method, shown in
two application areas. Bottleneck~1 cuts the number of rounds. An LLM agent turns
prior knowledge and past results into a Bayesian-optimization proposal, drawn
here for antibody bioprocess optimization. Bottleneck~2 cuts the cost per round.
Pre-trained surrogates predict an expensive measurement from cheap signals
(hardness to tensile strength, and X-ray with composition and CALPHAD to phase
fraction) for metal additive manufacturing, the lab we are building.}
\label{fig:arch}
\end{figure*}

\section{Agents Automate Everything but the Experiment}
\label{sec:intro}

Agentic AI-for-Science can automate much of the research workflow apart from real-world experiments, spanning ideation, planning, experimental design, analysis, and writing, as illustrated by the Virtual Lab, the AI Scientist, and Coscientist~\citep{swanson2025virtuallab,lu2024aiscientist,yamada2025aiscientistv2,boiko2023coscientist}. However, lab validation remains a physical constraint: a hypothesis is confirmed only when it is tested in real-world experiments. To mitigate this, as those upstream stages accelerate, the self-driving lab (SDL) has emerged as a promising way to reduce the time and cost of validation by automating repeated experiments and measurements~\citep{macleod2020self,szymanski2023autonomous}.

SDLs mainly focus on automating experiment execution, but a human still decides whether the proposed design of experiments (DOE) is feasible, useful, and scientifically informative. For an agentic SDL, that decision must be made by the agent itself, not by human intervention. We tackle two bottlenecks in this agent--bench loop: first, whether the agent can propose the next DOE by leveraging domain knowledge, feedback, and bench-side feasibility; and second, whether the agent can account for the time and resources needed to execute and measure each proposed experiment.

\section{Bottleneck 1: The Agentic DOE Loop}
\label{sec:bottleneck1}
In an SDL, the agent proposes a DOE, the lab executes it under physical constraints, and the returned results inform the agent's next proposal.
The bottleneck is that this agent--bench interaction cannot always produce better results.
Without combining domain priors, experimental feedback, and feasibility checks, the loop can proceed unproductively, spending rounds on low-value experiments.

Bayesian optimization (BO) formalizes sequential experimental design by updating a surrogate with each returned result and selecting the next DOE by balancing promising conditions against uncertain regions~\citep{frazier2018bo,balandat2020botorch}.
However, vanilla BO usually encodes expert knowledge and laboratory constraints only indirectly, for example through fixed bounds, hand-designed variables, or manually curated search spaces.
This motivates an agent-driven BO-DOE loop in which domain-specific expert knowledge, literature priors, and bench-side constraints actively shape each proposal before execution.
The agent suggests candidate regions from these priors, ranks them using past results, and removes infeasible designs before execution.

For example, in antibody process development, the agent proposes a small-scale bioreactor DOE, the lab measures titer and product quality, and the next DOE updates feeding,
pH, temperature, dissolved oxygen, and other culture conditions~\citep{warr2011microbioreactor,schiel2015bioreactor}.
These small-scale rounds are run before scale-up, because mini-bioreactors and scale-down systems screen conditions and predict larger manufacturing performance~\citep{warr2011microbioreactor,delouvroy2015ambr}.
For a pharmaceutical company, finishing this optimization quickly lets a new antibody program move into scaled production sooner.
In this setting, agentic DOE is expected to reduce trials-to-target by steering each round toward experimentally feasible and objective-relevant conditions.

We will compare against human-guided DOE, random search, grid search, and vanilla BO under the same experimental-loop budget, measuring trials-to-target, infeasible-proposal rate, and feedback sensitivity.

\section{Bottleneck 2: The Agentic Measurement Loop}
\label{sec:bottleneck2}

The AutoResearch concept introduced by Andrej Karpathy frames research as an iterative loop in which an agent proposes a change, executes it, evaluates the result, and keeps useful changes~\citep{karpathy2026autoresearch}.
This loop is straightforward in computational settings, where iterations can be repeated with low overhead.
Moving the same idea to a wet-lab SDL is harder, since each loop consumes lab time, materials, and measurement capacity.
If every candidate is sent to high-cost, high-resolution measurement, the loop becomes limited by measurement throughput rather than by the agent's ability to propose candidates~\citep{charbottleneck2024}.

To reduce this cost, we propose a cost-aware surrogate agent that predicts selected high-cost, high-resolution measurements from low-cost, low-resolution ones.
When the surrogate uncertainty is low, the agent uses the prediction from the low-cost measurement result; when the uncertainty is high, the agent requests the high-cost measurement~\citep{angelopoulos2021conformal}.
This allows the SDL to evaluate more candidates under the same budget by spending high-cost measurements only where they are most informative~\citep{costaware2026,coorch2024}.
The surrogate is not used to guess an expensive measurement from weak inputs; it is used only when the low-cost measurements contain enough information to estimate the target reliably~\citep{amlimit2025}.
When a single low-cost measurement is not informative enough, the agent combines multiple inputs so that their complementary information constrains the prediction.

For example, the metal additive-manufacturing SDL can provide two cases.
First, a quick hardness test provides a cheap proxy for tensile strength within one alloy family~\citep{pavlina2008}. This proxy is useful when hardness and strength change together, but it should not be trusted when tensile failure is dominated by hidden porosity that indentation cannot detect.
Second, phase fractions can be estimated from a combination of X-ray diffraction (XRD)~\citep{xrdphase2025}, elemental composition, and a CALPHAD phase-diagram calculation~\citep{phasefraction2024}, reducing reliance on slower electron-microscopy measurements such as SEM/EBSD. Each input alone may be insufficient, but together they constrain the target quantity.
Together, these cases illustrate the same principle: the agent uses low-cost, fast measurements when they provide enough information, and reserves high-cost characterization for candidates where the surrogate remains uncertain.

Following this design, we are building a metal-AM SDL that links directed energy deposition (DED) printing, sample cutting, heat treatment, and X-ray, optical, and electron characterization into one loop.
We will evaluate surrogate accuracy and uncertainty calibration on held-out measurements, on-target hits per characterization cost relative to measuring everything, and wall-clock time to reach a target specification.

\section{Conclusion}
\label{sec:conclusion}

The physical experiment is the rate-limiter for agentic scientific discovery. We focus on two places where an SDL still wastes time: too many experimental loops and a high cost for each loop. A prior-aware agentic DOE uses feedback to choose better next experiments, while a cost-aware surrogate agent uses low-cost measurements to avoid unnecessary high-cost measurements. The objective is ``reach the target faster, with fewer experiments within the budget''.

\bibliography{example_paper}
\bibliographystyle{icml2026}

\newpage
\appendix
\onecolumn

\section{Method Landscape and Positioning}
\label{app:landscape}

This appendix organises the references cited in the main text by methodological family and the question each addresses for our framework. The goal is to make explicit (i) where the proposed agent sits in the current LLM + Bayesian optimization (BO) landscape, (ii) which line of work the surrogate characterization is built on, and (iii) which self-driving lab (SDL) and additive-manufacturing (AM) precedents anchor the case study.

\subsection{LLM-augmented Bayesian optimization (Bottleneck 1)}
\label{app:llmbo}

Recent work has placed LLMs at different points in the BO loop. We compare them on four axes that matter for our framing, split across two tables to keep each readable: \Cref{tab:llmbo-prior} covers \emph{where the LLM enters the loop} and \emph{how prior knowledge is represented}; \Cref{tab:llmbo-feedback} covers \emph{how feedback is handled} and the \emph{empirical domain} where each system has been validated.

\begin{table*}[h]
\centering
\footnotesize
\caption{LLM placement and prior representation
(\S\ref{app:llmbo}, part 1 of 2). Last row is the agent proposed in this
work.}
\label{tab:llmbo-prior}
\begin{tabular}{@{}l p{5.5cm} p{6.5cm}@{}}
\toprule
\textbf{Method} & \textbf{LLM role in loop} & \textbf{Prior representation} \\
\midrule
LLAMBO~\citep{liu2024llambo}             & Warm-start, sample proposer, surrogate augmentation & Natural-language conditioning \\
BO-ICL~\citep{ramos2024boicl}            & LLM-as-regression-surrogate (in-context)            & Prompt + experimental history \\
LGBO~\citep{yuan2026lgbo}                & Per-round region/point preference                   & Region/point + confidence \\
ChemBOMAS~\citep{han2025chembomas}       & Pre-loop search-space decomposition                 & Chemistry RAG + LLaMA fine-tune \\
Coscientist~\citep{boiko2023coscientist} & Full plan-and-execute agent                          & Free-form chat reasoning \\
ChemCrow~\citep{bran2024chemcrow}        & Tool-augmented LLM                                  & Tool outputs + chat history \\
\citet{gupta2025arewethereyet}           & \emph{Diagnostic only} (no new system)              & --- \\
\midrule
\textbf{This work}                       & DOE proposer \emph{and} domain selector             & Physical-context block $\mathcal{P}$ + history $\mathcal{H}_t$ + literature \\
\bottomrule
\end{tabular}
\end{table*}

\begin{table*}[h]
\centering
\footnotesize
\caption{Feedback handling and empirical setting
(\S\ref{app:llmbo}, part 2 of 2). ``Random-label test'' refers to the
diagnostic introduced by~\citet{gupta2025arewethereyet}: an LLM agent passes
the test only if its selections change when feedback labels are randomly
permuted.}
\label{tab:llmbo-feedback}
\begin{tabular}{@{}l p{6.5cm} p{5.5cm}@{}}
\toprule
\textbf{Method} & \textbf{Feedback handling} & \textbf{Empirical domain} \\
\midrule
LLAMBO~\citep{liu2024llambo}             & In-context history (no posterior update on the LLM) & Hyperparameter tuning \\
BO-ICL~\citep{ramos2024boicl}            & Re-prompt each round with full history              & Catalysis (RWGS, OCM benchmarks) \\
LGBO~\citep{yuan2026lgbo}                & Mean-shift on GP from LLM region (forward-only)     & Physics, chemistry, biology, materials \\
ChemBOMAS~\citep{han2025chembomas}       & Tree refinement (no closed loop on outcomes)         & Buchwald and Suzuki couplings + wet-lab \\
Coscientist~\citep{boiko2023coscientist} & Natural-language summaries fed to next prompt        & Pd-catalysed cross-couplings \\
ChemCrow~\citep{bran2024chemcrow}        & Tool-result ingestion via chat                       & Synthesis, retrosynthesis, repellent design \\
\citet{gupta2025arewethereyet}           & Reveals LLM agents are insensitive to feedback (random-label test) & Gene perturbation, molecules \\
\midrule
\textbf{This work}                       & Surrogate-mediated; verifier-gated; both prior and outcome routed through the GP/neural surrogate, not the LLM alone & Antibody bioprocess (case study) \\
\bottomrule
\end{tabular}
\end{table*}

Two observations frame our position. First, LGBO~\citep{yuan2026lgbo} is the closest published system: it integrates the LLM in every round (rather than once at warm-start), and gives the only theoretical guarantee in the table (bounded performance loss when the LLM is misaligned). It is, however, restricted to a single target domain and emits region/point preferences only.
Second,~\citet{gupta2025arewethereyet} have shown empirically that off-the-shelf LLM experimental-design agents (BioDiscoveryAgent, LLAMBO) frequently \emph{fail} to use feedback: replacing true outcomes with permuted labels does not change selections. Our framework does not bypass this failure mode by trusting the LLM more. Instead, we route both prior and feedback through the GP/neural surrogate and a feasibility verifier so feedback effects are measurable.

\subsection{Surrogate Characterization and Multi-fidelity Modelling (Bottleneck 2)}
\label{app:mf}

Treating a cheap, fast measurement as a low-fidelity view of an expensive one is the multi-fidelity recipe. Co-kriging fuses the two through a learned relationship~\citep{kennedy2000bayesian}, multi-fidelity optimization adds adaptive sampling between fidelity levels~\citep{forrester2007multifidelity}, and recent guidance characterizes when a cheap source is worth using~\citep{mfbo2025}.

Our contribution is to make the cheap source a measurement modality rather than a coarser simulation, and to keep its use honest. The closest work is microViT~\citep{microvit2025}, which predicts a property from a frozen foundation backbone with no task-specific training. We place such a surrogate inside the agent loop and gate it by calibrated uncertainty, so the agent decides per candidate when a real measurement is worth its cost. This is different from co-orchestration~\citep{coorch2024}, which schedules the next instrument rather than replacing it, and from low-cost hardware laboratories, which cut the price of the machine rather than the cost of the measurement.

\subsection{Self-driving Labs and Characterization Precedents (Case Study)}
\label{app:sdlam}

Closed-loop self-driving labs were popularised in chemistry and materials science~\citep{macleod2020self,szymanski2023autonomous,bran2024chemcrow}. For metal AM, AIDED~\citep{shang2025aided} couples a genetic algorithm with machine-learning surrogates to invert process parameters in laser directed-energy deposition in about an hour per round. It is a Bottleneck~1 system that tightens the on-rig loop on the expensive metal target alone. On the characterization side, deep models now read an X-ray diffraction pattern into phase fractions about two orders of magnitude faster than Rietveld refinement~\citep{xrdphase2025}, and phase fractions also follow from composition and temperature through CALPHAD-trained surrogates~\citep{phasefraction2024}. Bottleneck~2 places these single-shot predictors under one uncertainty-gated agent, so a cheap prediction is trusted only when its calibrated uncertainty is low, and a real measurement is triggered otherwise.

\subsection{Agentic Upstream Stack (Motivation)}
\label{app:agentic}

The motivating works in the introduction, namely the Virtual Lab~\citep{swanson2025virtuallab}, the AI Scientist~\citep{lu2024aiscientist,yamada2025aiscientistv2}, Coscientist~\citep{boiko2023coscientist}, and ChemCrow~\citep{bran2024chemcrow}, are not direct competitors to the framework proposed here. Each demonstrates that a sufficiently capable agent can run the upstream pipeline
(ideation, planning, paper writing, or chemistry-tool orchestration) at or near human productivity. They are cited because they collectively establish that the rate-limiter has shifted: the physical experiment, not the agent turn, now dominates the schedule for high-value low-throughput science. Bottleneck~1 and Bottleneck~2 are responses to that shift, not extensions of those systems.

\end{document}